\documentclass{article} 
\usepackage{MLDD_workshop_2023,times}

\usepackage{hyperref}
\usepackage{url}

\newcommand{\bs}{\boldsymbol}

\newcommand{\emrow}{\rowcolor[rgb]{0.937,0.937,0.937}}
\newcommand{\rulesep}{\unskip\ \textcolor{gray} \vrule \ }

\newcommand{\eg}{{\em e.g.}}
\newcommand{\ie}{{\em i.e.}}

\newcommand{\x}{{\bs{x}}}
\renewcommand{\xi}{\x^{i}}

\newcommand{\z}{{\bs z}}

\newcommand{\params}{{\bs \theta}}

\newcommand{\pjoint}{\mathcal{P}}

\newcommand{\pdec}{p}
\newcommand{\penc}{q}

\newcommand{\Mdec}{\pdec_{\params}}
\newcommand{\Menc}{\penc_{\params}}

\newcommand{\AEloss}{\mathcal{L}_\text{AE}}
\newcommand{\VAEloss}{\mathcal{L}_\text{VAE}}

\newcommand{\MIMloss}{\mathcal{L}_\text{MIM}}
\newcommand{\AMIMloss}{\mathcal{L}_\text{A-MIM}}
\newcommand{\EAMIMloss}{\hat{\mathcal{L}}_\text{A-MIM}}
\newcommand{\DKL}[2]{\mathcal{D}_\text{KL}\left(#1\,\|\, #2\right)}
\newcommand{\E}[2]{\mathbb{E}_{#1}\left[#2\right]}

\title{Improving Small Molecule Generation using Mutual Information Machine}

\author{Danny Reidenbach\textsuperscript{\textasteriskcentered, 1, 2}, Micha Livne\thanks{equal contributions} \textsuperscript{\,, 2}, Rajesh K. Ilango\textsuperscript{2}, Michelle Gill\textsuperscript{2} \& Johnny Israeli\textsuperscript{2} \\
\textsuperscript{1}UC Berkeley, \textsuperscript{2}NVIDIA\\
\texttt{dreidenbach@berkeley.edu}, ~ \texttt{mlivne@nvidia.com}
}

\usepackage{times}  
\usepackage{helvet}  
\usepackage{courier}  
\usepackage{graphicx} 
\usepackage{natbib}  
\usepackage{caption} 
\usepackage{wrapfig}

\usepackage{algorithm}
\usepackage{algorithmic}

\usepackage{newfloat}
\usepackage{listings}
\DeclareCaptionStyle{ruled}{labelfont=normalfont,labelsep=colon,strut=off} 
\floatstyle{ruled}
\newfloat{listing}{tb}{lst}{}
\floatname{listing}{Listing}

\usepackage{natbib}
\usepackage{amsmath,amsfonts,bm}
\usepackage{url}
\usepackage{arydshln}
\usepackage{xcolor}
\usepackage{colortbl}
\usepackage{multirow}
\usepackage{subcaption}
\usepackage{nicefrac}
\usepackage{placeins}

\iclrfinalcopy 
\begin{document}

\maketitle

\begin{abstract}
We address the task of controlled generation of small molecules, which entails finding novel molecules with desired properties under certain constraints. 
Here we introduce MolMIM, a probabilistic auto-encoder for small molecule drug discovery that learns an informative and clustered latent space.
MolMIM is trained with Mutual Information Machine (MIM) learning and provides a fixed-size representation of variable-length SMILES strings.
Since encoder-decoder models can learn representations with ``holes'' of invalid samples, here we propose a novel extension to the MIM training procedure which promotes a dense latent space and allows the model to sample valid molecules from random perturbations of latent codes.
We provide a thorough comparison of MolMIM to several variable-size and fixed-size encoder-decoder models, demonstrating MolMIM's superior generation as measured in terms of validity, uniqueness, and novelty.
We then utilize CMA-ES, a naive black-box, and gradient-free search algorithm, over MolMIM's latent space for the task of property-guided molecule optimization.
We achieve state-of-the-art results in several constrained single-property optimization tasks and show competitive results in the challenging task of multi-objective optimization.
We attribute the strong results to the structure of MolMIM's learned representation which promotes the clustering of similar molecules in the latent space, whereas CMA-ES is often used as a baseline optimization method. 
We also demonstrate MolMIM to be favorable in a compute-limited regime.
\end{abstract}

\section{Introduction}

\begin{wrapfigure}{R}{0.5\textwidth}
 \centering
 \begin{minipage}{.5\textwidth}
 \includegraphics[width=1.0\columnwidth]{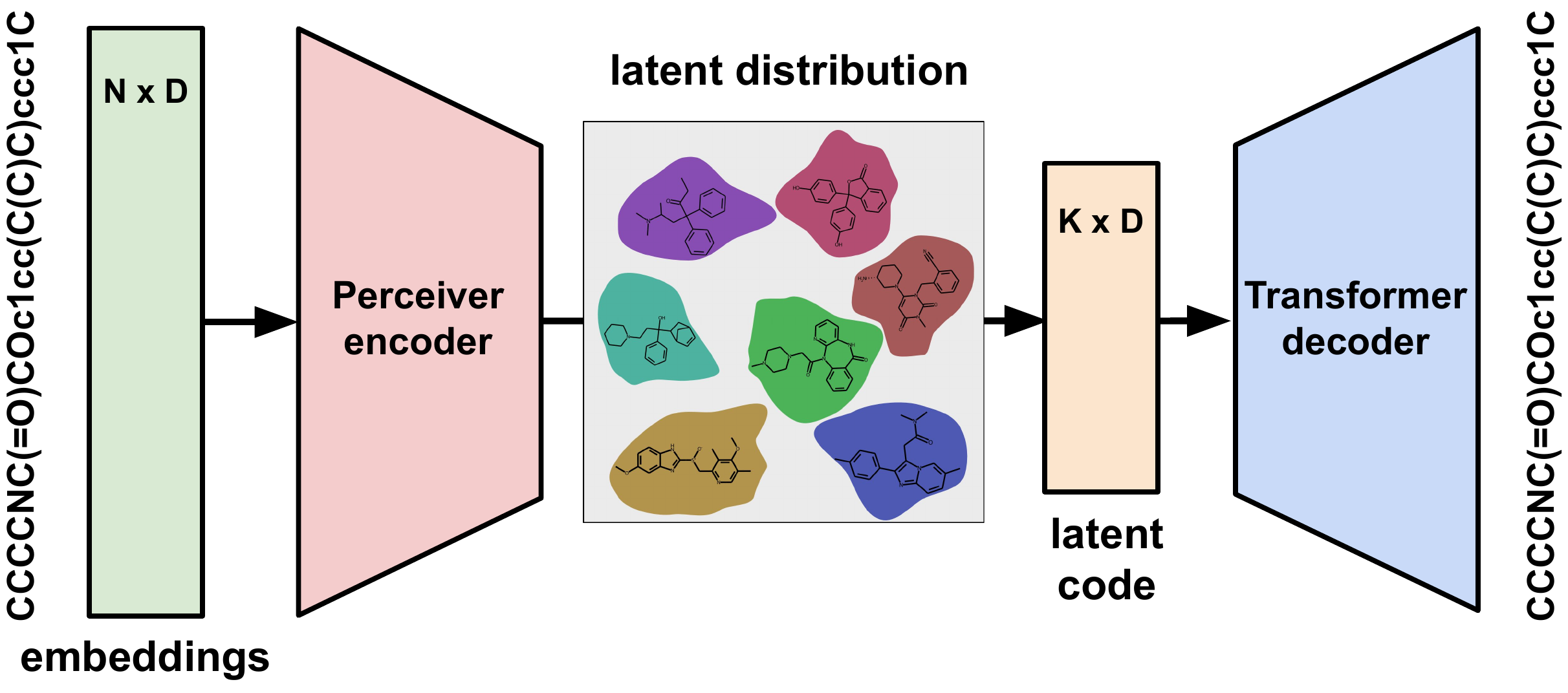}
 \caption{
 MolMIM is a Mutual Information Machine (MIM), a probabilistic encoder-decoder model with a fixed-size representation that learns an informative latent distribution clustered around similar molecules.
 N is the variable tokens number, D is the embeddings dimension, K is the fixed hidden length. Tokens are mapped to learnable embeddings.
}
  \label{fig:sampling} ~
  \end{minipage}
\end{wrapfigure}


The lead optimization stage of the drug discovery process is time consuming, labor intensive, and has a high rate of failure, requiring as much as three years and hundreds of millions of dollars for a single drug. This stage is focused on optimizing candidate molecules using the design-make-test cycle, in which scientists design new molecules based on available assay information, synthesize these molecules, and then test them in new assays. 
The risk and associated cost of this process makes it a high-value automation task in the drug discovery pipeline \citep{Kim2020-cu, Engkvistch13}. Successful solutions to the automated design of small molecules can be directly applied to related challenges associated with other biological modalities, such as protein design \citep{https://doi.org/10.48550/arxiv.2205.04259} and optimization of guide RNA sequence for CRISPR systems \citep{Chuai2018}.

Controlled generation of small molecules entails finding a new molecule with certain properties and under some constraints (e.g., similarity to a reference molecule, \citet{Vamathevan2019}). Often these molecules are represented in a text-based encoding format called SMILES \citep{doi:10.1021/ci00057a005}. Efficient search of the space of molecules is a challenging problem due to the high dimensional and sparse nature of samples, where valid molecules are sparse given all possible combinations of legal characters in SMILES. 

A common search method relies on genetic algorithms to modify a molecule's SMILES representation using heuristics. Examples would be random mutations and hand-crafted rules \citep{Sliwoski334,Mahmood2021}. The complex and high dimensional search space often leads to low sampling efficiency of such methods. In addition, the search is often based on \emph{ad hoc} rules which require input from experts.

An alternative approach to automate this process with deep learning is to project the discrete molecules into a continuous space, wherein generation becomes sampling from a continuous space, and exploration becomes a manipulation of continuous vectors \citep{doi:10.1021/acscentsci.7b00572,qmo,C8SC04175J}. Here, we follow that approach, focusing on learning a dense representation space that allows efficient sampling and exploration (Fig. \ref{fig:sampling}). 

In this paper we propose a novel probabilistic encoder-decoder model named \textit{MolMIM}, a Mutual Information Machine \citep{2019arXiv191003175L} for large-scale SMILES pretraining.
Depicted in Fig.\ \ref{fig:sampling}, we leverage a Perceiver encoder with a Transformer Decoder to learn an informative and clustered fixed-sized latent space which is promoted by Mutual Information Machine (MIM) learning.
We empirically show that MolMIM clusters similar molecules together, and in contrast to prior work we emphasize that the latent organization occurs implicitly without integrating chemical property information into the training procedure  (\eg, \citet{C8SC04175J}). 
We also compare MolMIM with additional architectures with and without a bottleneck, and with other latent regularization methods, showing the effect on the ability to sample novel molecules, and on the performance of constrained molecular optimization in the latent space.

\textbf{Main Contributions:} We present MolMIM, a novel latent variable model that is trained in an unsupervised manner over a large-scale SMILES dataset. 
We show how MolMIM is capable of efficiently sampling unique and novel molecules with only slight perturbations of the latent code. 
We show that MolMIM learns an informative fixed-sized latent space in which chemically similar molecules are clustered together without explicit chemical property information.
We demonstrate how  a naive latent optimization strategy over MolMIM's latent space outperforms competing complex optimization methods. 
We set multiple state-of-the-art results, demonstrating the effectiveness of MolMIM in single and multi-property optimization, and under a limited compute regime.

\section{Formulation} \label{sec:formulation}

MolMIM expands upon prior SMILES-based modeling techniques by leveraging a Transformer architecture to learn an informative fixed-size latent space using MIM learning. 
We propose to utilize a Perceiver \citep{pmlr-v139-jaegle21a} encoder architecture which outputs a fixed-size representation, where molecules of various lengths are mapped into a latent space, allowing us to learn latent variable models, and in particular, MIM and VAE, the Variational Auto-Encoder \citep{Kingma2014AutoEncodingVB}.

\begin{figure}[t]
\centering
\begin{minipage}[t]{0.95\columnwidth}
\begin{algorithm}[H]
    \small
    \caption{Learning parameters $\params$ of MolMIM}
    \label{algo:molmim}
    \begin{algorithmic}[1]
        \REQUIRE Samples from dataset $\pjoint(\x)$
        \WHILE{not converged}
        \STATE $\sigma \sim \mathcal{U}(0,1]$
        \STATE $D \gets \{ \x_j, \z_j \sim \Menc(\z|\x,\sigma)\pjoint(\x) \}_{j=1}^{N}$
        \STATE $\EAMIMloss \left( \params ; D \right) = -\frac{1}{N}\! \sum_{i=1}^{N}\! \big( ~ \log \Mdec(\x_i | \z_i) + \frac{1}{2} \left( \log \Menc(\z_i | \x_i,\sigma) + \log \pjoint(\z_i) \right) ~ \big)$
        \STATE $\Delta \params \propto -\nabla_{\params}  \EAMIMloss \left( \params ; D \right)$
        \COMMENT{\textcolor{gray}{\textit{Gradient computed through sampling using reparameterization}}}
        \ENDWHILE
    \end{algorithmic}
\end{algorithm}
\end{minipage}
\end{figure}

MIM is a learning framework for a latent variable model which promotes informative and clustered latent codes. We use MIM to avoid the main caveat of VAE, a phenomenon called posterior collapse where the learned encoding distribution closely matches the prior, and the latent codes carry little information \citep{razavi2018preventing}. 
Posterior collapse leads to poor reconstruction accuracy, where the learned model performs well as a sampler, but allows little control over the generated molecule.
We further provide the relation between MIM, VAE, and Auto-Encoder (AE) in Appendix \ref{sec:model-def}.

Here we use A-MIM learning \citet{2019arXiv191003175L}, a MIM variant\footnote{Compared to MIM, A-MIM avoids sampling from a slow auto-regressive decoder during training} that minimizes the following loss per observation,
\begin{equation}
    \AMIMloss (\params) = \E{\x \sim \pjoint(\x),\z \sim \Menc(\z|\x)}{ \log \Mdec(\x | \z) + \log \pjoint(\z) + \log \Menc(\z | \x) + \log \Menc(\x)} \label{ea:mim-loss}
\end{equation}
where the expectation of $\z$ is taken over samples $\z \sim \Menc(\z | \x)$, similar to VAE, $\pjoint(\x)$ is the given data distribution (\ie, the SMILES dataset), and $\pjoint(\z)$ is the prior over the latent variable.
Minimizing $\AMIMloss(\params)$ trains a model with a consistent
encoder-decoder, high mutual information, and low marginal entropy.
In practice, we define the marginal probabilities to be the expectation over the conditional distributions \citep{https://doi.org/10.48550/arxiv.2003.02645},
and the final loss that is minimized is given in Alg. \ref{algo:molmim}.

Unlike VAE, MIM does not suffer from posterior collapse. However, like VAE, it might learn a posterior marginal distribution that does not match the prior, leaving ``holes'' where corresponding latent codes are decoded into invalid samples \citep{pmlr-v148-li21a, https://doi.org/10.48550/arxiv.2003.02645}.
Here, we introduce a novel yet simple extension to A-MIM learning which empirically mitigates the sampling of invalid molecules. 

During training, we sample the posterior's standard deviation,
\begin{equation}
    \Menc(\z | \x, \sigma) \equiv \mathcal{N}\big(\z | \mu_\params(\x, \sigma), \sigma \big)
    \label{ea:mim-posterior-sample}
\end{equation}
where $\sigma \sim \mathcal{U}(0,1]$ is sampled uniformly, and where the posterior is conditioned on the sampled $\sigma$ via linear mapping which is prepended to the input embedding.
This, in effect, is training a model that can accommodate different levels of uncertainty, and is encouraged to learn a dense latent space that supports sampling with little to no ``holes''. 
During inference time, we can adaptively choose the desired uncertainty in our latent space according to particular downstream tasks.
We provide the full training procedure in Algorithm \ref{algo:molmim}, where $\mathcal{P}(\z)$ is a Normal distribution. 

In order to understand why such a training procedure would work, it is first important to remember that MIM learning promotes high mutual information (MI) between the observation $\x$ and the latent code $\z$.
The maximization of MI prevents the decoder from ignoring the conditioning latent code. Conditioning the encoder on the variance allows the latent code to carry the uncertainty to the decoder, which then learns to support accurate reconstruction when a small variance is provided. Although MIM has been applied to text before as seen in SentenceMIM \citep{https://doi.org/10.48550/arxiv.2003.02645}, MolMIM is the first instance of MIM applied to molecules.
\section{Related Work}

The challenges associated with the task of molecular optimization have been thoroughly researched, and here we described models that relate to the scope of this work.
SMILES VAE \citep{doi:10.1021/acscentsci.7b00572} learns physical chemical information by jointly training a molecular property predictor from the continuous latent codes.
CDDD \citep{C8SC04175J} similarly leverages character-level SMILES encoding but is trained with an additional chemical property regression loss to structure the resulting latent space.
To take advantage of the underlying graphical structure of molecules, JT-VAE, AtomG2G/HierVAE and VJTNN/VJTNN+GAN \citep{jt-vae, g2g, vjtnn} leverage graph neural networks to learn a VAE for property optimization and graph translation.

VSeq2Seq \citep{vseq2seq}, Chemformer \citep{Irwin_2022}, and more recently MoLFormer-XL\citep{molformerxl} are sequence-to-sequence models which are based on RNN \citep{Rumelhart1986LearningRB}, BART \citep{lewis-etal-2020-bart}, and Transformer with relative positional embeddings, respectively. 
VSeq2Seq and Chemformer can be considered the precursor of MolMIM. Both MolMIM and MoLFormer-XL apply large-scale pretraining, but we focus on generative tasks instead of property prediction. 



There have been several examples of deep learning methods for molecular optimization and design. 
While each relies on fundamentally different design choices, all attempt to generate molecules under strict constraints such as improved properties, target binding affinity, solubility, or reduced toxicity.
Previous approaches such as MolDQN \citep{MolDQN}, GCPN \citep{GCPN}, REINVENT \citep{Olivecrona2017}, GVAE-RL/RationaleRL \cite{barzilay-multi-obj}, and FaST \citep{FAST} leverage reinforcement learning (RL) for single and multi-property molecule optimization. These methods use various molecule representations, including fragments and motifs to build expressive generative models. 

Outside of RL, there have been major successes in using fragment-based approaches in MMPA \citep{mmpa} and MARS \citep{mars}, query-based guided search QMO \citep{qmo}, and molecular fingerprint to SMILES translation in DESMILES \citep{desmiles} for property optimization.
Several methods use flow-based models, including MoFlow \citep{moflow}, MolGrow \citep{molgrow}, GraphAF \citep{graphaf}, and GraphDF \citep{graphdf} for efficient molecule graph generation. 
Genetic algorithms such as GA \citep{GA} and JANUS \citep{JANUS} and diffusion models like CDGS \citep{CDGS} have also shown success in property-guided molecular optimization.
\section{Experiments}
In what follows, we evaluate MolMIM against several SMILES-based architectures on small molecule generation tasks to better understand the effect of different model components on the generative performance. 
We discuss \textit{MegaMolBART} \citep{megamolbart_ngc}, a BART model trained on SMILES data; 
\textit{PerBART}, a Perceiver BART, where we replace the Transformer encoder with a fixed-size output Perceiver encoder;
\textit{MolVAE}, a VAE which shares the architecture with PerBART and has two additional linear layers to project the Perceiver encoder output to a mean and variance of the posterior; 
and \textit{MolMIM} which shares the same architecture as MolVAE with an additional linear layer to project the Gaussian posterior's standard deviation into the input embedding.
For a detailed description of the training, including architecture, data, and optimization procedure please see supplementary Appendix \ref{sec:training-details}.

We remind the reader that both VAE and MIM are latent variable models that are implemented with almost identical encoder-decoder architectures but possess subtle differences in their loss functions. 
VAE tends to learn a smooth representation (\ie, roughly matching the Gaussian prior) but suffers from posterior collapse; \ie, manifested via uninformative latent codes and poor reconstruction, see discussion in \citet{Bowman2016GeneratingSF}. 
In contrast, MIM learns an informative and clustered representation and does not suffer from posterior collapse by design \citep{2019arXiv191003175L}.

\subsection{Sampling Quality} \label{sec:sampling}

\begin{table*}[htb]
\renewcommand{\arraystretch}{1.1}
\resizebox{\textwidth}{!}{%
\begin{tabular}{l||cc|c|cccc|c|cc}
\textbf{Model} & K & \textbf{Latent Dim.} & \textbf{Eff. Nov.(\%)} & \textbf{Validity(\%)} & \textbf{Unique(\%)} & \textbf{Non Id.(\%)} & \textbf{Novelty(\%)} & $\bs \sigma$ & \textbf{Test Time} & \textbf{Batch} \\ 
\hline
\hline
{MMB} & - & variable & 51.1 & 75 & 84.8 & 74.4 & 93.1 & 1.2 & 8.7 hours~ & 100 \textdagger \\ 
{PerBART} & 4 & 2048 & 59.1 & 71.8 & 94.9 & 88.4 & 94.3 & 0.7 & 38 min~ & 500 \\ 
{MolVAE} & 4 & 2048 & 93.9 & 95.7 & \textbf{100} & \textbf{100} & 98.1 & 1.2 & 63 min & 500 \\
\emrow {MolMIM}  & 1 & \textbf{512} & \textbf{94.2} & \textbf{98.7} & \textbf{100} & 99.9 & 95.5 & 1.42 & \textbf{30 min} & 500 \\ 
\hline
CDDD & 1 & \textbf{512} & 82.2 & 84.5 & 98.9 & 98 & \textbf{99.4} & 1.2 & 12 hours~ & 1 \\ 
\end{tabular}%
}
\caption{Molecule sampling quality was evaluated with 20,000 molecules randomly selected from the test set, where 10 samples were acquired per molecule.
MMB stands for MegaMolBART. $K$ is the hidden length. $\sigma$ is the optimal scale of Gaussian random noise used in sampling. \textdagger batch size constrained by memory.
Top models are developed herein.}
\label{table:sampling}
\end{table*}

Our motivation is to be able to generate useful (\eg, novel, see discussion below) molecules for a variety of optimization tasks from our pre-trained models.
To do so, we perturb the latent code of a given molecule by adding Gaussian noise.
In more detail, sampling entails
\begin{equation}
    \x' \sim \Mdec(\x | \z' = \z + \epsilon)
\end{equation}
where the latent code $\z = \E{z}{\Menc(\z | \x)}$ is taken to be the posterior mean, and $\epsilon\sim \mathcal{N}\big(\mu=0, \sigma \big)$ is noise sampled from a Gaussian with a given standard deviation $\sigma$.
In the case of MegaMolBART and PerBART, which have deterministic encoders, we define a posterior as a Dirac delta around the encoder output.

In what follows, we compare MolMIM to various models, measuring their sampling performance according to the following common metrics \citep{doi:10.1021/acs.jcim.8b00839}:
\textit{Validity} is the percentage of generated molecules that are valid SMILES;
\textit{Uniqueness} is the percentage of generated valid molecules that are unique; and 
\textit{Novelty} is the percentage of generated valid and unique molecules that are not present in the training data.\footnote{Validity determined by RDKit: http://www.rdkit.org}

In addition, we add the following metrics:
\textit{Non-Identicality} is the percentage of valid molecules that are not identical to the input; 
\textit{Effective Novelty} is the percentage of generated molecules that are valid, non-identical, unique, and novel.
Effective novelty was created to provide a single metric that measures the percentage of ``useful'' molecules when sampling, combining all other metrics in a practical manner.
We point the reader to Appendix \ref{sec:sampling-metrics} for an additional discussion.

Table \ref{table:sampling} shows the results of the various sampling metrics, as defined above, for the best performing models (\ie, following hyper-parameters search). 
We performed a grid search over the hidden length $K \in \{1,2,4,8,16\}$ (\ie, for Perceiver-based models only), and the sampling noise scale $\sigma \in (0,2]$ (in 0.1 increments, for all models) in order to maximize the effective novelty.
The results were computed with 10 sampled molecules per input molecule.
We randomly sampled 20,000 input molecules from the ZINC-15 \citep{doi:10.1021/acs.jcim.5b00559} test set at the optimal noise scale per model, and report the average sampling metrics.

We conclude, based on Table \ref{table:sampling}, that the overall sampling quality is improved by the bottleneck architecture of PerBART, and further improved by regularization over the latent codes (\ie, MolMIM, MolVAE).
Importantly, MolMIM's best effective novelty was achieved using the smallest number of latent dimensions, which is beneficial for sample-based optimization in the latent space.
The usefulness of effective novelty is demonstrated by the CDDD model, which has the best novelty out of all of the benchmarked models, but its effective novelty is 12\% lower than MolMIM. 
We note that MolMIM was trained on nearly 10x the data compared to CDDD. Each model's novelty calculation was based on their respective training sets.


\subsection{Latent Space Structure} \label{sec:latent-structure}

\begin{wrapfigure}{R}{0.5\textwidth}
 \centering
 \includegraphics[width=0.5\columnwidth]{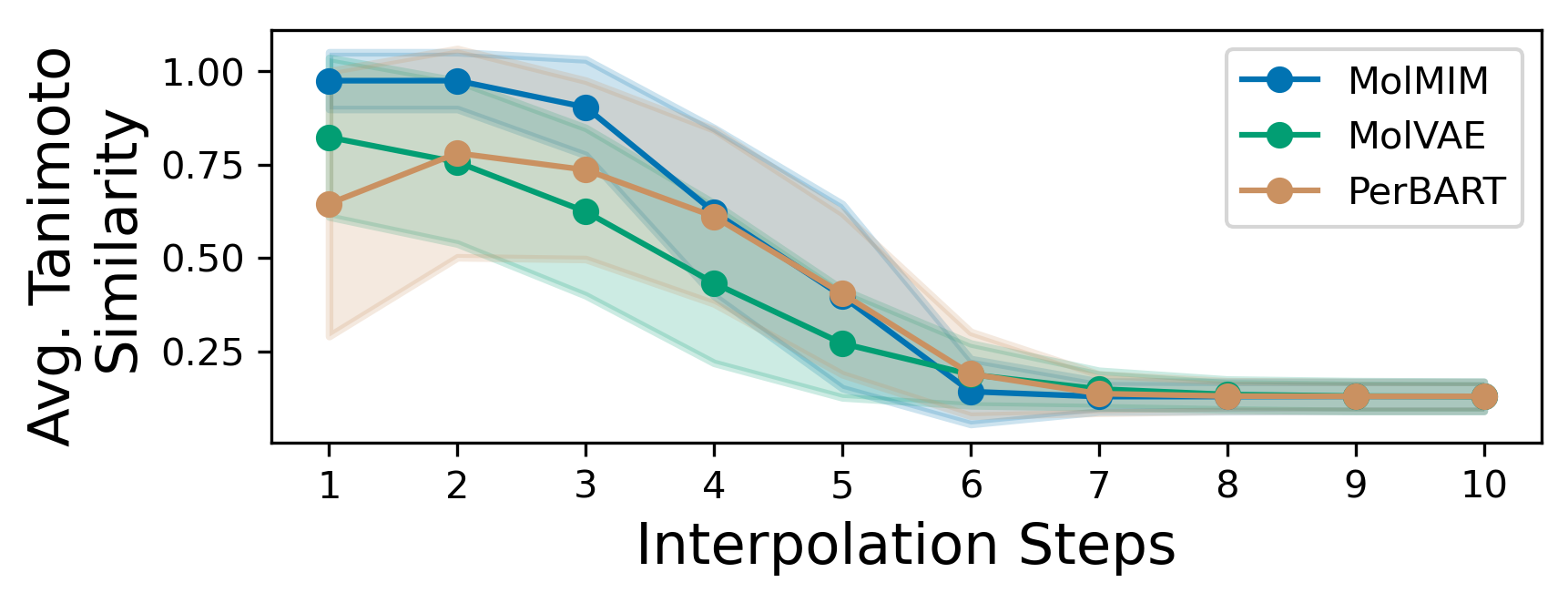}
 \caption{Average Tanimoto similarity (y-axis) between input and non-identical interpolated molecules (x-axis is interpolation step).
 Colored areas indicate standard deviation. MolMIM clusters similar molecules and has the smallest variance in early steps.}
 \label{fig:similarity-interpolation}
\end{wrapfigure}

\begin{figure*}
     \centering
     \hfill
     \begin{subfigure}[b]{0.19\textwidth}
         \centering
         \includegraphics[width=\textwidth]{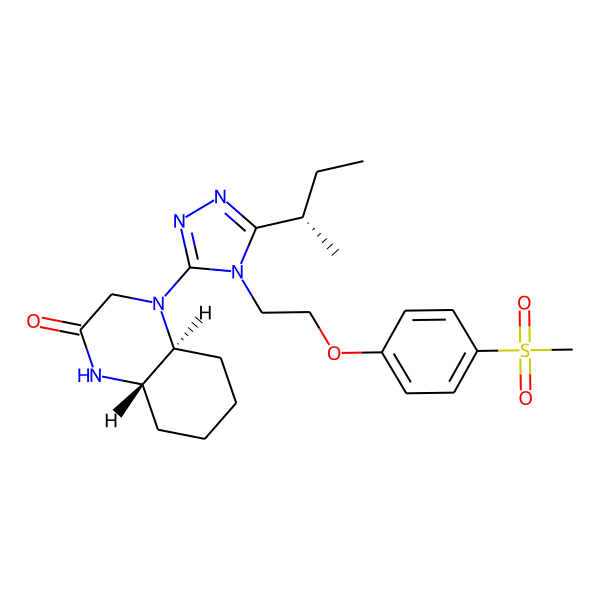}
         \caption{Initial}
         \label{fig:mol-original1}
     \end{subfigure}
     \begin{subfigure}[b]{0.19\textwidth}
         \centering
         \includegraphics[width=\textwidth]{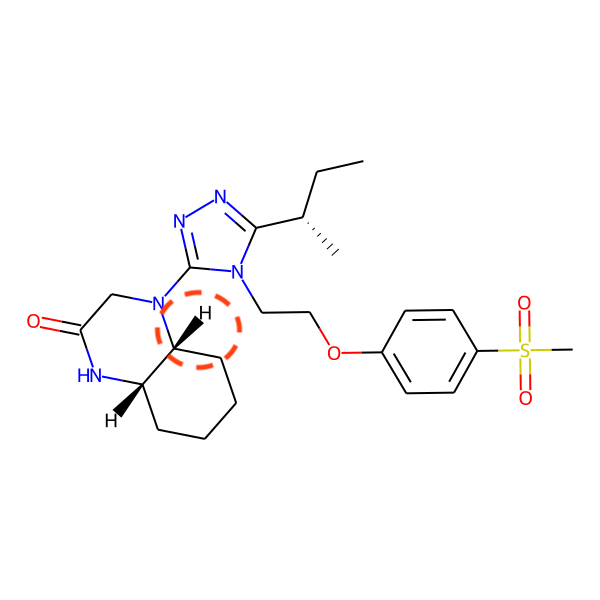}
         \caption{Perturbed}
         \label{fig:mol-perturbed1}
     \end{subfigure}
     \rulesep
     \begin{subfigure}[b]{0.19\textwidth}
         \centering
         \includegraphics[width=\textwidth]{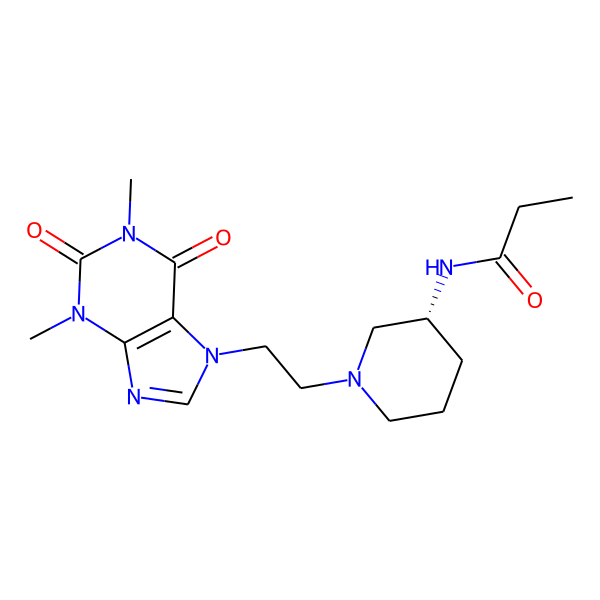}
         \caption{Initial}
         \label{fig:mol-original2}
     \end{subfigure}
     \begin{subfigure}[b]{0.19\textwidth}
         \centering
         \includegraphics[width=\textwidth]{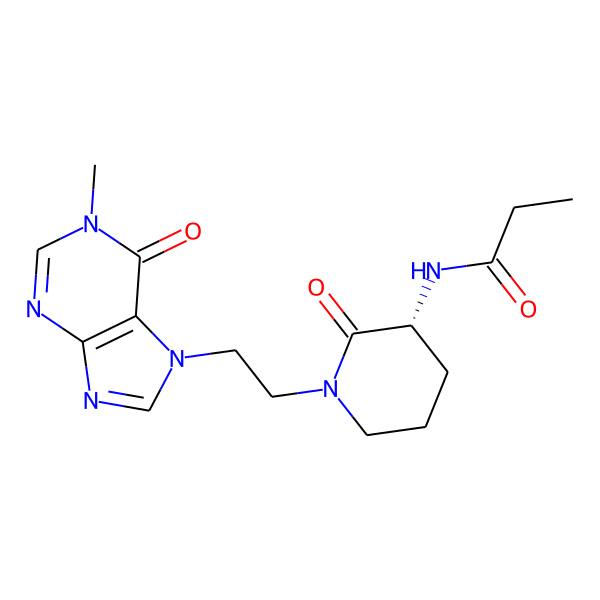}
         \caption{Perturbed}
         \label{fig:mol-perturbed2}
     \end{subfigure}
     \begin{subfigure}[b]{0.19\textwidth}
         \centering
         \includegraphics[width=\textwidth]{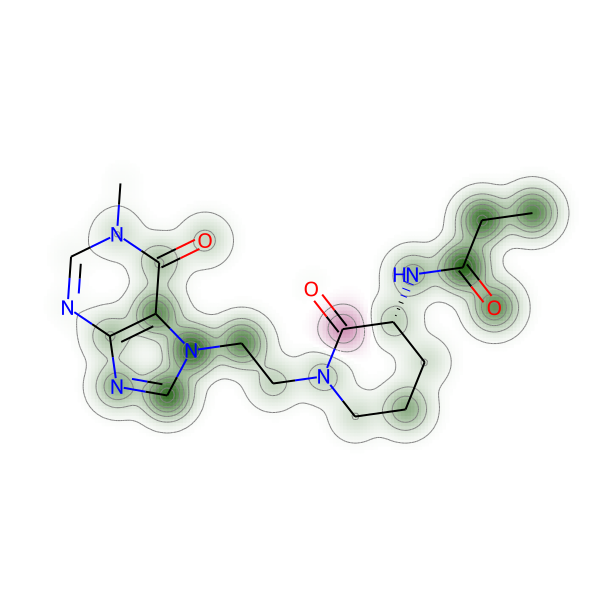}
         \caption{Similarity Map}
         \label{fig:mol-map2}
     \end{subfigure}
     \hfill
 \caption{MolMIM's fine-grained control over molecule generation. (a-b) Small perturbations lead to changes in chirality only (circled in dashed red). 
 (c-d) Bigger changes allow the substitution of a single atom (red background in similarity map (e)).
 The similarity map depicts green background for chemically similar structures and red background for modified structures.}
 \label{fig:mol-fine-grain-control}
\end{figure*}

In addition to the ability to sample valid, unique, and novel molecules, we would like to train a model with a learned latent structure that will lend itself to the optimization of molecular properties.
In the context of unsupervised learning, the structure can relate to the similarity of observations, or intrinsic properties of latent codes (\eg, smoothness).
We hypothesize that a latent space where latent codes of similar molecules are clustered together will allow for fine-grained control while searching for molecules with target properties.

In what follows we explore the clustering of molecules in the latent space using pair-wise interpolations of 1,000 test set molecules.
This allows us to better understand the qualitative differences in the spatial structures of the respective latent spaces of each model.
An interpolation entails projecting two molecules onto the latent space by taking the latent codes to be the respective mean of the posterior for each molecule. 
Following the method used by \citet{doi:10.1021/acscentsci.7b00572}, we then linearly interpolate between the latent codes over 10 equidistant steps, and for each interpolated latent code we decode a corresponding molecule. 
We then compute the Morgan fingerprint \citep{doi:10.1021/ci100050t} with 2048 bits and radius 2, resulting in a bit vector of hashed molecular features for each molecule. Finally, the average Tanimoto similarity \citep{doi:10.1021/ci100050t} is calculated between the Morgan fingerprints of the initial and all non-identical interpolated molecules.  

Fig. \ref{fig:similarity-interpolation} shows that MolVAE's smooth latent space results in a gradual similarity decline, whereas MolMIM contains regions of high similarity followed by a sharper decline. 
MolVAE and PerBART show lower average similarity for interpolation step 1. For MolVAE, it is due to poor reconstruction in the absence of noise. 
For PerBART, the less ordered structure of its latent space leads to a quick divergence when small amounts of noise are added.

In contrast, MolMIM maintains near perfect similarity for steps 1 and 2, while producing non-identical molecules. 
MolMIM also demonstrates a significantly lower variance in steps 1-3, making it a reliable sampler when considering similarity.
This is an interesting result as MolMIM is not explicitly trained with Tanimoto similarity information\footnote{Tanimoto similarity cannot be calculated from SMILES, requiring conversion to Morgan Fingerprints}.
We conclude from Fig. \ref{fig:similarity-interpolation} that the latent structure of MolMIM is clustered by meaningful chemical similarity.

Fig. \ref{fig:mol-fine-grain-control} depicts qualitative examples of the fine-grained control supported by MolMIM. Fig. \ref{fig:mol-fine-grain-control}(a-b) demonstrate how small
perturbations can lead to minimal changes, such as inversions of a chiral center of a single molecule (circled in dashed red).
Fig. \ref{fig:mol-fine-grain-control}(c-e) depict more significant changes where the similarity map highlights in red a single-atom change. 
The ability to generate molecules of even more diversity is discussed next. 
\subsection{Small Molecule Optimization} \label{sec:controlled-generation}

In what follows we use CMA-ES \citep{Hansen06thecma}, a greedy, gradient-free (\ie, 0\textsuperscript{th} order optimization), evolutionary search algorithm that maximizes a black-box reward function for small molecule optimization. 
CMA-ES is often used as an optimization baseline (\eg, \citet{yang2021learning}), and does not scale well to high-dimensional problems.
We apply CMA-ES directly to the latent space to generate proposed latent solutions thus expanding on the sampling technique discussed in Sec. \ref{sec:sampling}. 
These putative solutions are then greedily decoded to generate molecules. 
The molecules are the required inputs to the reward function, which is comprised of molecular property oracle functions.
We chose CMA-ES as it is a simple algorithm and weak baseline for novel optimization methods thus allowing us to isolate the optimization success of MolMIM to its latent structure.
We believe that using more powerful latent optimization techniques, such as RL, could further improve the results.
We used TDC \citep{TDC} oracle functions for all chemical property values in this section.
Here we demonstrate how such a naive algorithm can deliver state-of-the-art results given an informative and clustered latent space.


\subsubsection{Single Property Optimization}

\begin{wraptable}{R}{0.5\textwidth}
{%
\begin{center}
\renewcommand{\arraystretch}{1.05}
\resizebox{0.5\columnwidth}{!}{
\begin{tabular}{l||c|cc} 
 & \textbf{QED (\%)} & \multicolumn{2}{c}{\textbf{Penalized logP} }  \\ 
\textbf{Task} & $\delta=0.4$ & $\delta=0.6$  & $\delta=0.4$  \\ 
\hline
\hline
JT-VAE & 8.8 & 0.28 $\pm$ 0.79 & 1.03 $\pm$ 1.39 \\ 
GCPN & 9.4 & 0.79 $\pm$ 0.63 & 2.49 $\pm$ 1.30 \\ 
MolDQN & - & 1.86 $\pm$ 1.21 & 3.37 $\pm$ 1.62 \\
MMPA & 32.9 & - & - \\ 
VSeq2Seq & 58.5 & 2.33 $\pm$ 1.17  & 3.37 $\pm$ 1.75 \\ 
VJTNN+GAN & 60.6  & - & - \\ 
VJTNN & -  & 2.33 $\pm$ 1.24 & 3.55 $\pm$ 1.67 \\ 
MoFlow & - & 2.10 $\pm$ 2.86 & 4.71 $\pm$ 4.55 \\ 
GA & -  & 3.44 $\pm$ 1.09 & 5.93 $\pm$ 1.41 \\
AtomG2G & 73.6 & - & - \\ 
HierG2G & 76.9 & - & - \\ 
DESMILES & 77.8 & - & - \\ 
QMO & 92.8 & 3.73 $\pm$ 2.85 & 7.71 $\pm$ 5.65 \\
MolGrow & - & 4.06 $\pm$ 5.61 & 8.34 $\pm$ 6.85 \\ 
GraphAF & - & 4.98 $\pm$ 6.49 & 8.21 $\pm$ 6.51 \\ 
GraphDF & - & 4.51 $\pm$ 5.80 & 9.19 $\pm$ 6.43 \\
CDGS & - & 5.10 $\pm$ 5.80 & 9.56 $\pm$ 6.33 \\
FaST & - & 8.98 $\pm$ 6.31 & 18.09 $\pm$ 8.72 \\
\hline
\emrow {MolMIM} & \textbf{94.6} & \textbf{7.60 $\pm$ 23.62} & \textbf{28.45 $\pm$ 54.67} \\
{MolMIM}\textdagger &  & 4.57 $\pm$ 3.87 & 9.44 $\pm$ 4.12 \\
\end{tabular}%
}
\end{center}%
}
\caption{QED and penalized logP optimization under minimal Tanimoto similarity constraint $\delta$. 
QED results show success percentage and penalized logP show the mean and standard deviation of the improvement in value.
\textdagger~ limits logP solutions to improvement $\leq$ 20. 
Bottom models are developed herein. 
For logP, 97.25\% and 79.13\% of the molecules saw an increase in score for $\delta \in {0.4, 0.6}$ respectively, 100\% were within the target $\delta$, and 0\% saw a decrease in score.
}
\label{table:singple-prop}
\end{wraptable}

In what follows we explore the optimization of a single chemical property under a Tanimoto similarity constraint $\delta$, which limits the allowed distance from an input molecule.
Specifically, we target Quantitative Estimate of Drug-likeness (QED, \citet{QED}) and penalized logP \citep{jt-vae}. 

We use 800 molecules with QED $\in [0.7,0.8]$ \citep{vjtnn}, and 800 molecules with low penalized logP scores \citep{jt-vae} as the starting points for our respective optimizations. 
For the QED task, the success rate is defined as the percentage of generated molecules with QED $\geq 0.9$ while maintaining at least a $\delta = 0.4$ Tanimoto similarity to the respective input \citep{vjtnn}. 
For the penalized logP task, we report the mean and standard deviation of penalized logP improvement under a similarity constraint for $\delta \in \{0.4,0.6\}$ \citep{jt-vae}.
We provide the reward functions and additional information in Appendix \ref{sec:app-mole-opt}.

In both experiments, we follow a query budget of 50,000 oracle calls per input molecule, following \citet{qmo}. 
We have found that dividing each optimization into 50 restarts with 1,000 CMA-ES iterations\footnote{We tested 100, 300, 400, 800, 1000, and 1600 iterations}, and a population size of 20 to yield the best results.
We omit MegaMolBART from the next experiments since it is unclear how to utilize CMA-ES in a variable-size representation. 

Table \ref{table:singple-prop} shows that MolMIM yields new state-of-the-art results in both QED and penalized logP optimization tasks.
We also point the reader to the large value of the standard deviation for the logP experiment, in which MolMIM exploits a known flaw in the logP oracle function where generating molecules with long saturated carbon chains yield high logP values \citep{RENZ201955}. 

\begin{wrapfigure}{R}{0.5\textwidth}
 \centering
 \includegraphics[width=0.5\columnwidth]{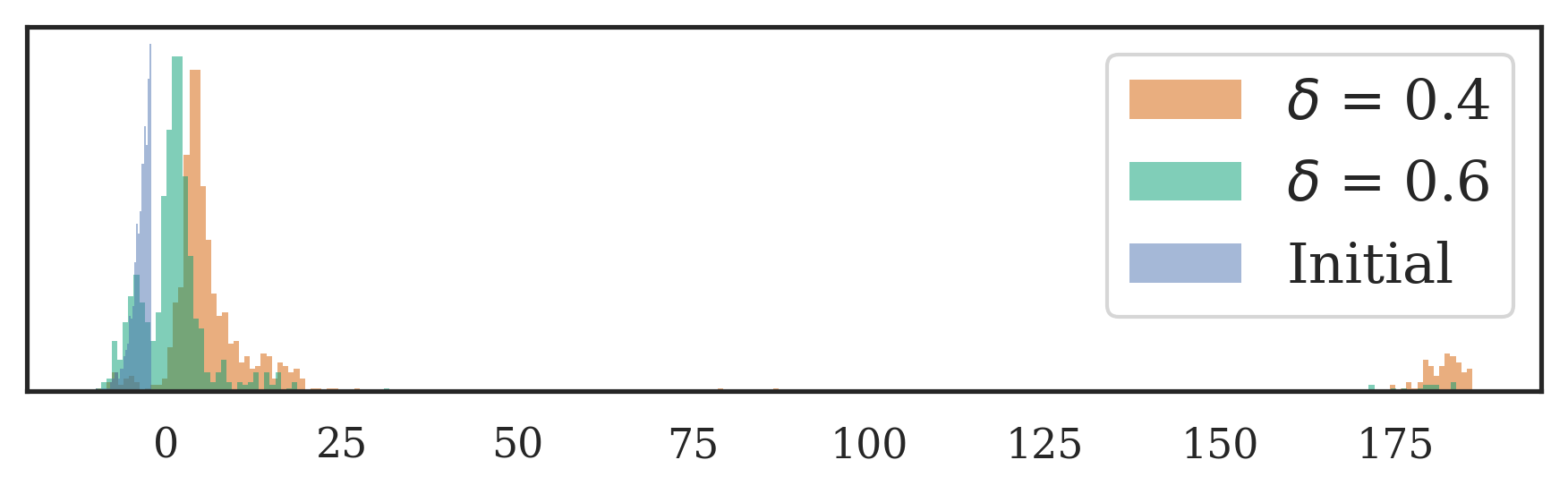}
 \caption{Penalized logP distribution of initial molecules (blue) and optimized molecules under increasingly stringent Tanimoto similarity constraints of 0.4 and 0.6 (orange and green, respectively).
 The right-hand side mode represents a common bias in the pen. logP oracle function.}
 \label{fig:logp}
\end{wrapfigure}

Fig. \ref{fig:logp} clearly shows a bimodal distribution with the right-hand side mode depicting the logP exploitation described above.
As these saturated molecules are not useful in practice \citep{RENZ201955}, we also provide results of MolMIM\textdagger\ in Table \ref{table:singple-prop}. 
Here we limit the maximal penalized logP improvement to 20 during the optimization procedure, removing the problematic mode.
By doing so we obtain a significantly smaller standard deviation while still maintaining a high mean value and 100\% success rate. We stress that the structure of MolMIM's latent space allows a simple CMA-ES to outperform CDGS which relied on a complex diffusion procedure and ODE solvers. Only FaST and MolMIM\textdagger\ can achieve a standard deviation less than its mean for the more difficult similarity constraint $\delta = 0.6$. 

As a final single property experiment, we point the reader to the Appendix \ref{sec:compute-limited} Table. \ref{table:compute-limited} where we use CMA-ES with a limited iteration budget to optimize over each latent space. MolMIM consistently provides superior results over all other models, including CDDD, especially at the lower limit of 100 iterations (roughly 2x performance increase).

\subsubsection{Multi-Objective Property Optimization}
\begin{wraptable}{R}{0.5\textwidth}
{%
\begin{center}
\renewcommand{\arraystretch}{1.1}
\resizebox{0.5\columnwidth}{!}{%
\begin{tabular}{l||ccc} 
 & \multicolumn{3}{c}{\textbf{GSK3$\beta$ + JNK3 + QED + SA} }  \\ 
\textbf{Model} & Success (\%) & Novelty (\%)  & Diversity  \\ 
\hline
\hline
JT-VAE & 1.3 & - & - \\ 
GVAE-RL & 2.1 & - & - \\ 
GCPN & 4.0 & - & - \\ 
REINVENT & 47.9 & - & - \\ 
RationaleRL & 74.8 & 56.1 & 0.621  \\ 
MARS & 92.3 & 82.4 & 0.719  \\ 
JANUS & \textbf{100} & 32.6 & \textbf{.821} \\
FaST & \textbf{100} & \textbf{100} & 0.716 \\
\hline
\emrow {MolMIM (R)} & 97.5 & 71.1 & 0.791 \\
{MolMIM (A)} & 96.6 & 63.3 & 0.807 \\
{MolMIM (E)} & 98.3 & 55.1 & 0.767 \\
{MolMIM (E)}\textdagger & 99.2 & 54.8 & 0.772 \\
\end{tabular}%
}
\end{center}%
}
\caption{Multi-objective molecule optimization. (R) random initialization, (A) promising precursor initialization, (E) initialization with exemplars. \textdagger~ results are based on additional restarts.
Bottom models are developed herein.
}
\label{table:multi-prop}
\end{wraptable}

Due to the fact that the unbounded optimization of penalized logP contradicts Lipinski's rule of 5 \citep{lipinski}, we chose a more challenging setting that better represents the complexity of real-world drug discovery \citep{coley-survey, FAST}.
The task is defined over multiple objectives, where a randomly chosen molecule is jointly optimized for $\text{QED} \ge 0.6$, $\text{SA} \le 4.0$, $\text{JNK3} \ge 0.5$, and $\text{GSK3}\beta \ge 0.5$. 
We follow the same procedure and model usage as defined by \citet{barzilay-multi-obj} for predicting inhibition of JNK3 and GSK3$\beta$.

We consider three variants of initial molecule selection: (R) random start - 2,000 initial molecules are randomly sampled from the ZINC-15 test set; 
(A) approximate start - 551 initial molecules that satisfy all three conditions $\text{QED} \in [0.25, 0.4), \text{JNK3} \in [0.25, 0.35), \text{GSK3}\beta \in [0.25, 0.35)$; (E) exemplar-based - 741 initial molecules that already satisfy the required success conditions. In the case of (E), we also force the optimized molecule to have at most 0.4 Tanimoto similarity to the initial molecule.
The multi-property experiments entailed 28 restarts of 1500 iterations with a CMA-ES population size of 20. We provide the reward functions in Appendix \ref{sec:app-mole-opt}.

Table \ref{table:multi-prop} shows the results measured by success rate, novelty, and diversity, as defined by \citet{barzilay-multi-obj}. 
Here, novelty is the percentage of generated molecules with Tanimoto similarity less than 0.4 compared to the nearest neighbor in the ChEMBL training set \citep{Olivecrona2017}.
Diversity is defined as the pairwise Tanimoto similarity over Morgan fingerprints between all generated molecules $1 -  \frac{2}{n(n-1)} \sum_{\x \ne \x'} TanSim(\x,\x')$ where $\x$ is sampled without replacement.

Compared to JANUS, a genetic algorithm, MolMIM achieves competitive success rate and diversity while significantly improving novelty \footnote{Novelty here does not measure the percent of novel generated molecules as done in Sec. \ref{sec:sampling}}. MolMIM also achieves better diversity than FaST a recent molecular fragment-based optimization method that uses a novelty/diversity-aware RL policy. We note that MolMIM achieves competitive results without RL or other complex optimization methods and can later be combined with such in future research.

We also present in Table \ref{table:multi-prop} results for MolMIM (E)\textdagger, where we launched additional restarts, reaching 49 in total.
In such a case we present a very high success rate at the expense of novelty. 
We added MolMIM (E) to mimic more realistic drug development where we generate a new molecular structure while maintaining the desired properties. 
MolMIM (A) targets drug development use cases where, starting from a promising precursor, we want to generate a successful related molecule. MolMIM (E) increases the difficulty of the optimization task and now not only do we have to meet four property thresholds we also introduce a similarity maximum constraint.

In summary, we present three variants for multi-objective optimization. MolMIM demonstrates competitive results using a simple CMA-ES algorithm.
We attribute the success of MolMIM to the informative and clustered latent space which proved to be useful for various optimization tasks. 

\section{Conclusions}
In this paper, we present a novel probabilistic auto-encoder for small molecules called MolMIM.
Trained with Mutual Information Machine (MIM) learning, the model learns an informative and clustered latent space and samples novel molecules with high probability.
We utilize MolMIM to set multiple state-of-the-art results in single and multiple property optimization tasks through the incorporation of a simple search algorithm, CMA-ES.
Importantly, any successful solution to the small molecule optimization problem can be applied to other biological modalities such as proteins, RNA, and DNA. 
Future research directions will include improvement of the latent space search, replacing CMAE-ES with a more informed search algorithm.
In addition, the utilization of graphical molecular representations for training MolMIM might yield further improvement.

\bibliography{paper}
\bibliographystyle{iclr2023_conference}

\appendix
\clearpage
\section{Supplementary Material}


\subsection{Formal Model Definitions} \label{sec:model-def}

Here we discuss the relation between auto-encoder (AE), VAE, and MIM. 
We explicitly show that MIM and VAE extends AE with a regularization term that promotes certain properties of the latent space, which are different for the two models.


\subsubsection{Denoising Auto-encoder}
MolMIM builds upon denoising auto-encoders \citep{Goodfellow-et-al-2016} where a corrupted input is encoded by an encoder into a latent code. 
The latent code is then used to reconstruct the original input by the decoder. 
More formally, we can describe auto-encoders (AE) in terms of encoding distribution $\Menc(\z | \x)$ and decoding distribution $\Mdec(\x | \z)$, 
where we opt here for a probabilistic view. A deterministic encoder can be viewed as a Dirac delta function around the predicted mean.
Given the encoder and decoder, the denoising AE (DAE) loss, per observation $\x$ can be expressed as,
\begin{equation}
    \AEloss (\params) = \E{\z \sim \Menc(\z | \tilde{\x})}{\log \Mdec(\x | \z)} \label{ea:ae-loss}
\end{equation}
where $\x \in \mathcal{V}^N$ for vocabulary $\mathcal{V}$, $\tilde{\x}$ is some kind of corruption or augmentation of $\x$, $\z \in \mathbb{R}^{H}$, $H$ is the hidden dimensions, and $\params$ is the union of all learnable parameters. Here we include the identity function in the set of augmentations, where $\tilde{\x} \equiv \x$. 


\subsubsection{Bottleneck Architectures}
A precursor to MolMIM is our baseline BART \citep{lewis-etal-2020-bart} model referred to as MegaMolBART, with data augmentation identical to Chemformer \citep{Irwin_2022}. 
BART is a transformer-based seq2seq model that learns a variable-size hidden representation $H = |\tilde{\x}| \times D$.
That is, the dimension of the hidden representation is equal to the number of tokens in the encoder input times the embedding dimension.
This makes sampling from the model especially challenging since the molecule length has to be sampled as well. 
In contrast, learning a fixed-size representation as typically done in denoising auto-encoders, where all molecules are mapped into the same space, makes sampling easier.

Thus, we propose to replace MegaMolBART's Transformer encoder with a fixed-sized output Perceiver encoder \citep{pmlr-v139-jaegle21a}.
Perceiver is an attention-based architecture that utilizes cross-attention to project a variable input onto a fixed-size output. 
More formally,  $\z \in \mathbb{R}^{H}$ for a pre-defined dimension $H$.


\subsubsection{Latent Variable Models}

A fixed-sized representation with a bottleneck architecture allows us to learn latent variable models (LVMs).
A popular LVM is a Variational Auto-Encoder (VAE) and was introduced by \citet{Kingma2014AutoEncodingVB}.
VAE training expands on the typical denoising AE with the following loss per observations, 
\begin{equation}
    \VAEloss (\params) = \AEloss (\params) + \DKL{\Menc(\z | \tilde{\x})}{\Mdec(\z)} \label{ea:vae-loss}
\end{equation}
where $\Mdec(\z)$ is the prior over the latent code, which is typically a Normal distribution as in our case.
The KL divergence term encourages smoothness in the latent space. 
We define our posterior $\Menc$ to be a Gaussian with a diagonal covariance matrix.
We sample $\z$ from the posterior using the reparametrization trick which leads to a low variance estimator of the gradient during training.

A main caveat of VAE is a phenomenon called posterior collapse where the learned encoding distribution is closely matching the prior, and the latent codes carry little information \citep{razavi2018preventing}.
Posterior collapse leads to poor reconstruction accuracy, where the learned model performs well as a sampler, but allows little control over the generated molecule.

In contrast to VAE, MIM learning entails minimizing the following loss per observation
\begin{equation}
    \MIMloss (\params) = \AEloss (\params) + \E{\Menc}{ \log \big( \Mdec(\z)\Menc(\z | \tilde{\x})\Menc(\x) \big)} \label{ea:mim-loss-ae}
\end{equation}
which promotes high mutual information between $\z$ and $\x$, and low marginal entropy in $\z$ (\ie, clustered representation).
As discussed in Sec. \ref{sec:formulation}, MolMIM was designed in response to the aforementioned flaws of variable-length transformer models and VAEs. To test our novel combination of bottleneck architecture and learning framework we compare MolMIM to MolVAE, and PerBART, which use the same Perceiver bottleneck architecture and general training procedure.

We note that the PerBART uses the same learning as a regular BART, which does not explicitly promotes any structure in the learned latent space.
We show empirically that BART learns a latent space with ``holes'', where sampling from a ``hole'' results in an invalid molecule.
PerBART was specifically created to exemplify how the added latent regularization of MolMIM and MolVAE promotes structure in the latent space. Doing so allows a greater understanding of how both the addition of a bottleneck as well as the form of latent regularization impact generative molecular tasks in different ways.


\subsection{Training Details} \label{sec:training-details}
\textbf{Dataset:}
All models were trained using a tranche of the ZINC-15 dataset \citep{doi:10.1021/acs.jcim.5b00559}, labeled as reactive and annotated, with molecular weight $\le$ 500Da and logP $\le$ 5. Of these molecules, 730M were selected at random and split into training, testing, and validation sets, with 723M molecules in the training set. We note that we do not explore the effect of model size, hyperparameters, and data on the models. Instead, we train all models on the same data using the same hyperparameters, focusing on the effect of the learning framework and the fixed-size bottleneck. For comparison, Chemformer was trained on 100M molecules from ZINC-15 \citep{doi:10.1021/acs.jcim.5b00559} -- 20X the size of the dataset used to train CDDD (72M from ZINC-15 and PubChem \citep{10.1093/nar/gky1033}). MoLFormer-XL was trained on 1.1 billion molecules from the PubChem and ZINC datasets.

\textbf{Data augmentation:}
Following \citet{Irwin_2022}, we used two augmentation methods: masking, and SMILES enumeration \citep{doi:10.1021/ci00057a005}. Masking is as described for the BART MLM denoising objective, with 10\% of the tokens being masked, and was only used during the training of MegaMolBART. 
In addition, MegaMolBART, PerBART, and MolVAE used SMILES enumeration where the encoder and decoder received different valid permutations of the input SMILES string.
MolMIM was the only model to see an increase in performance when both the encoder and decoder received the same input SMILES permutation,
simplifying the training procedure.

\textbf{Model details:}
We implemented all models with NeMo Megatron toolkit \citep{kuchaiev2019nemo}.
We used a RegEx tokenizer with 523 tokens \citep{bird2009natural}.
All models had 6 layers in the encoder and 6 layers in the decoder, with a hidden size of 512, 8 attention heads, and a feed-forward dimension of 2048.
The Perceiver-based models also required defining K, the hidden length, which relates to the hidden dimension by $H = K \times D$ where $H$ is the total hidden dimension, and $D$ is the model dimension (Fig. \ref{fig:sampling}).
MegaMolBART had $58.9 M$ parameters, PerBART had $64.6 M$, and MolVAE and MolMIM had $65.2 M$.
We used greedy decoding in all experiments.
We note that we trained MolVAE using the loss of $\beta$-VAE \citep{higgins2017betavae} where we scaled the KL divergence term with $\beta = \frac{1}{D}$ where $D$ is the hidden dimensions.

\textbf{Optimization:}
We use ADAM optimizer \citep{DBLP:journals/corr/KingmaB14} with a learning rate of 1.0, betas of 0.9 and 0.999, weight decay of 0.0, and an epsilon value of 1.0e-8.
We used Noam learning rate scheduler \citep{NIPS2017_3f5ee243} with a warm-up ratio of 0.008, and a minimum learning rate of 1e-5.
During training, we used a maximum sequence length of 512, dropout of 0.1, local batch size of 256, and global batch size of 16384.
All models were trained for 1,000,000 steps with fp16 precision for 40 hours on 4 nodes with 16 GPU/node (Tesla V100 32GB).
MolVAE was trained using $\beta$-VAE \citep{higgins2017betavae} with $\beta = \frac{1}{H}$ where $H$ is the number of hidden dimensions.
We have found this choice to provide a reasonable balance between the rate and distortion (see \citet{pmlr-v80-alemi18a} for details). It is important to note that MolMIM does not require the same $\beta$ hyperparameter tuning as done for VAE.

\subsection{Small Molecule Optimization} \label{sec:app-mole-opt}

In this section, we formulate the reward functions that were used in the small molecule optimization part of the main body.


\subsubsection{Single Property Optimization}
Quantitative Estimate of Druglikeness (QED) is a simple rule-based molecular property that measures drug-likeliness \citep{QED}.
Penalized logP \citep{jt-vae} is logP, which is the $\log_{10}$ of the octanol and water solute partition ratio and is a measure of hydrophobicity with larger values indicating increased hydrophobicity, minus the Synthetic Accessibility (SA) score \citep{sa-score}.

Formally, we define the following reward functions for our CMA-ES optimization:
\begin{align}
\mathcal{R}_{QED} & = \min(\frac{QED}{0.9},1) + \min(\frac{TanSim}{0.4},1)  \label{eq:Rqed}  \\ 
\mathcal{R}_{plogP} & = \frac{plogP}{20} + \min(\frac{TanSim}{\delta},1) \label{eq:Rlp} 
\end{align}
where $TanSim$ is the Tanimoto similarity, $\delta$ is the value of the similarity constraint, and $QED,\  plogP$ are the corresponding properties.
The loss scaling of each term was tuned manually over a few test runs. 


\subsubsection{Multi Property Property Optimization}

JNK3 is the inhibition of c-Jun N-terminal kinase-3. GSK3$\beta$ is the inhibition of glycogen synthase kinase-3 beta.

Formally, we show the reward function below,
\begin{align}
\mathcal{R}_{MP}^{R,A} & = \frac{QED + GSK3\beta}{2} + \frac{SA}{40} + JNK3 \label{eq:Rmpra} \\ 
\mathcal{R}_{MP}^{E} & =  \frac{1}{50} \mathcal{R}_{MP}^{R,A} + \min(1.0, \frac{1-TanSim}{1-0.4}) \label{eq:Rmpe}
\end{align}
where $TanSim$ is the Tanimoto similarity, and $QED,\ SA,\ JNK3,\ GSK3\beta$ are the corresponding properties.
The loss scaling of each term was tuned manually over a few test runs. 

\FloatBarrier


\subsection{Finding Optimal Noise Scale for Top Models}

\begin{figure}[htp]
 \centering
 \includegraphics[width=\columnwidth]{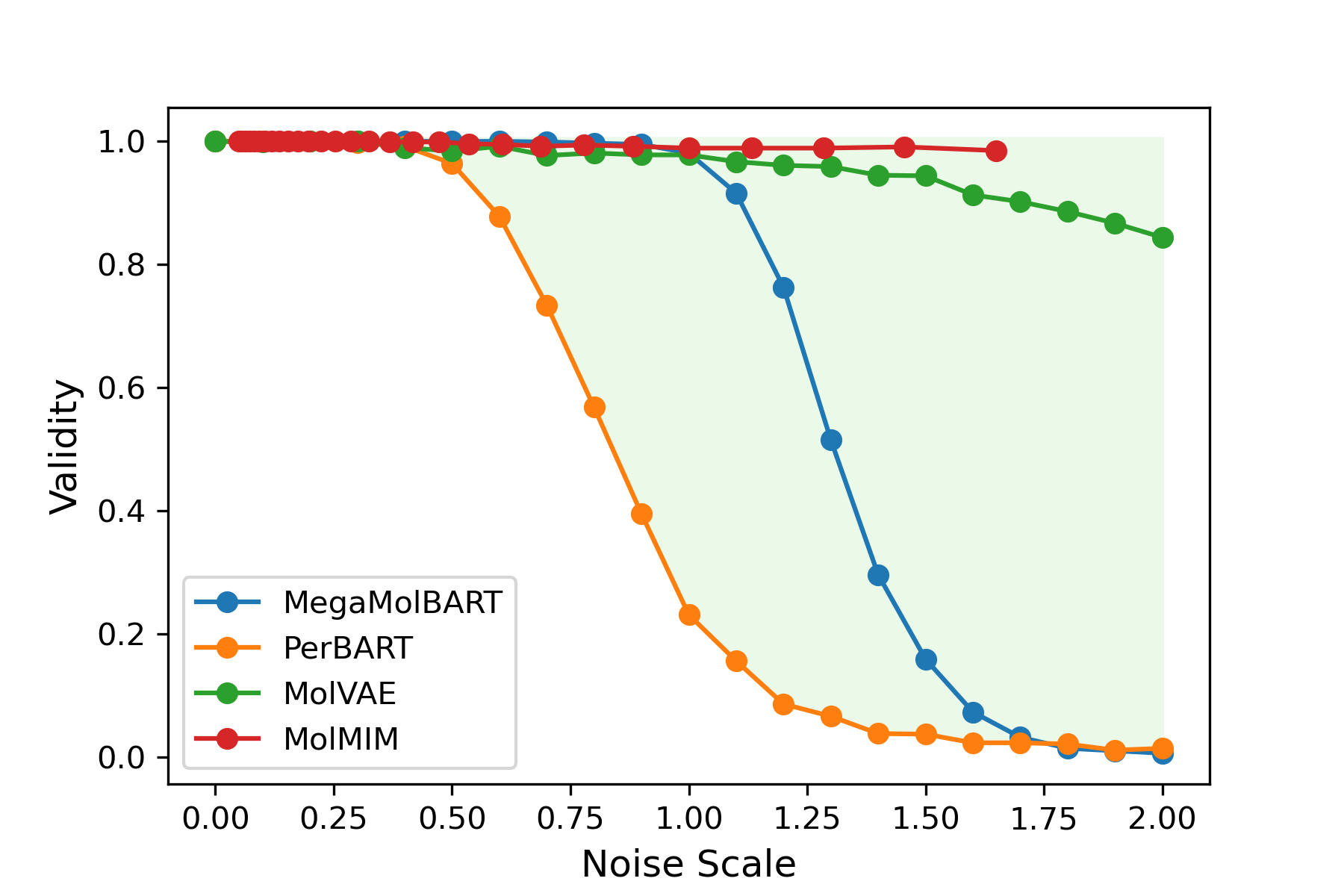}
 \caption{Validity as a function of noise scale. Note how latent variable models, MolVAE and MolMIM, are consistent while the others, MegaMolBART and PerBART, see a sharp decline at larger noise scales.}
  \label{fig:validity}
\end{figure}

\begin{figure}[htp]
 \centering
 \includegraphics[width=\columnwidth]{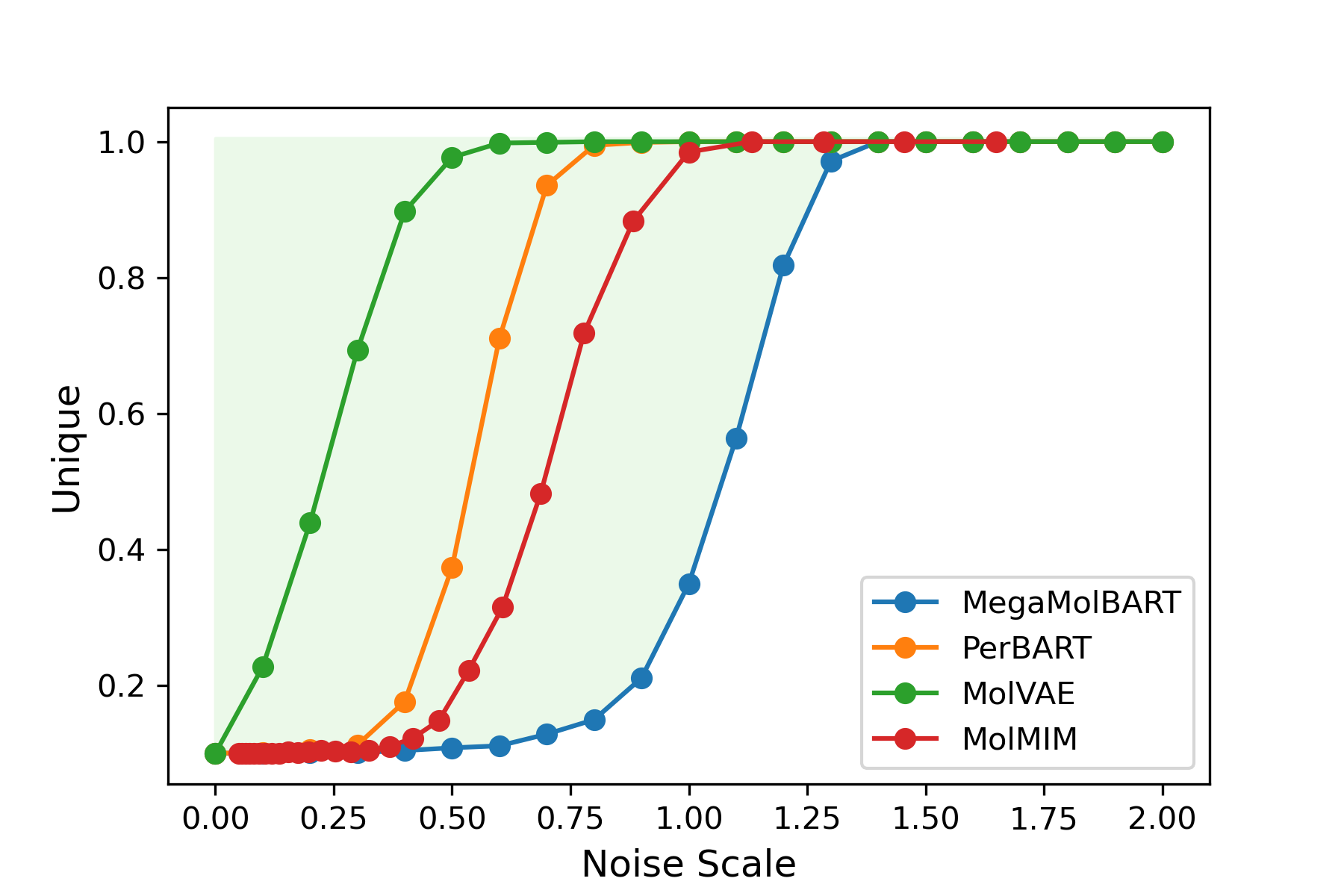}
 \caption{Uniqueness as a function of noise scale. Note MegaMolBART is a lower bound for the entire range of tested noise scales.}
  \label{fig:unique}
\end{figure}

\begin{figure}[htp]
 \centering
 \includegraphics[width=\columnwidth]{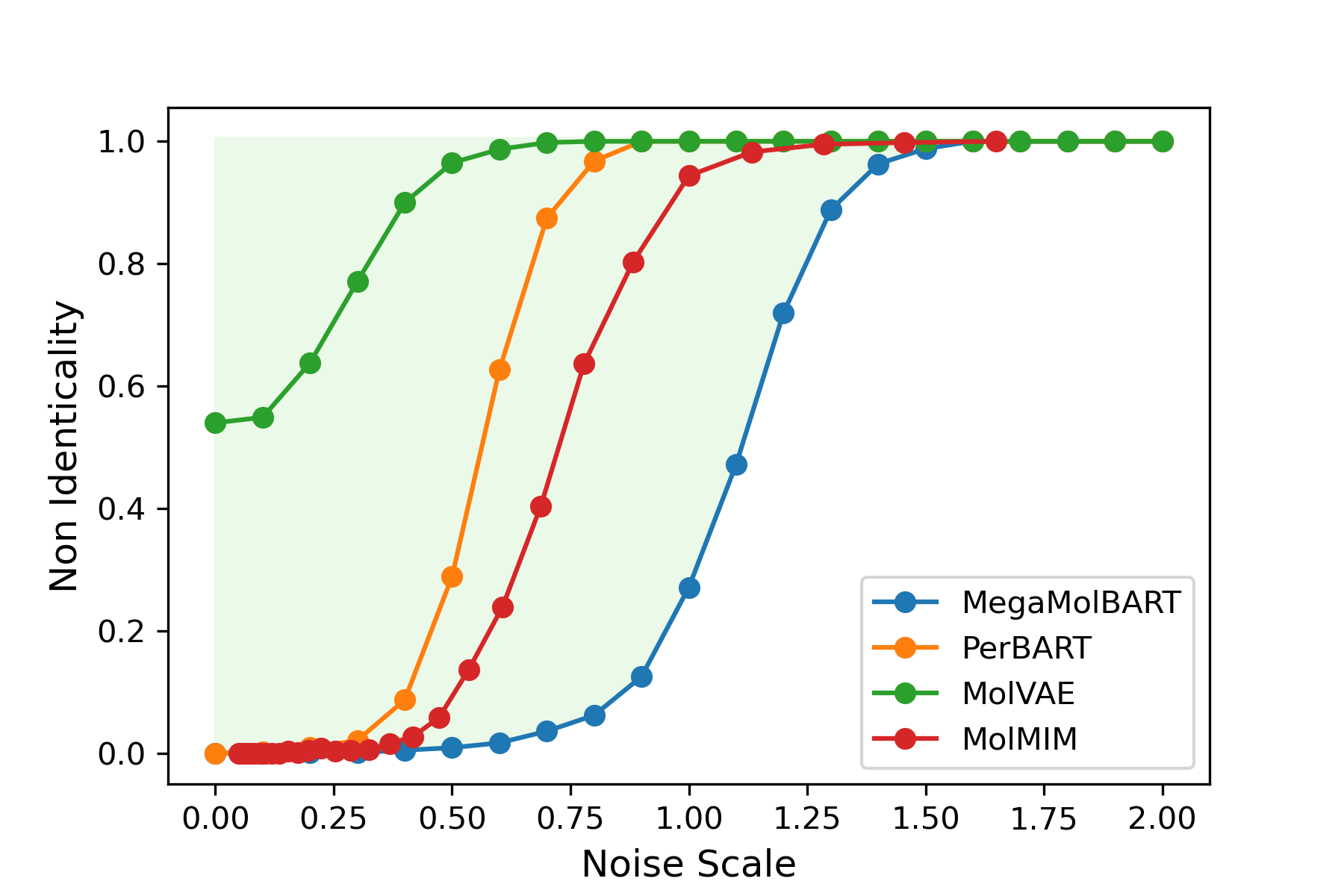}
 \caption{Non-Identicality as a function of noise scale. Note at a nose scale of 0, only MolVAE has a non zero non-identicality as a result of poor reconstruction (\ie, relates to posterior collapse).}
  \label{fig:non-iden}
\end{figure}

\begin{figure}[htp]
 \centering
 \includegraphics[width=\columnwidth]{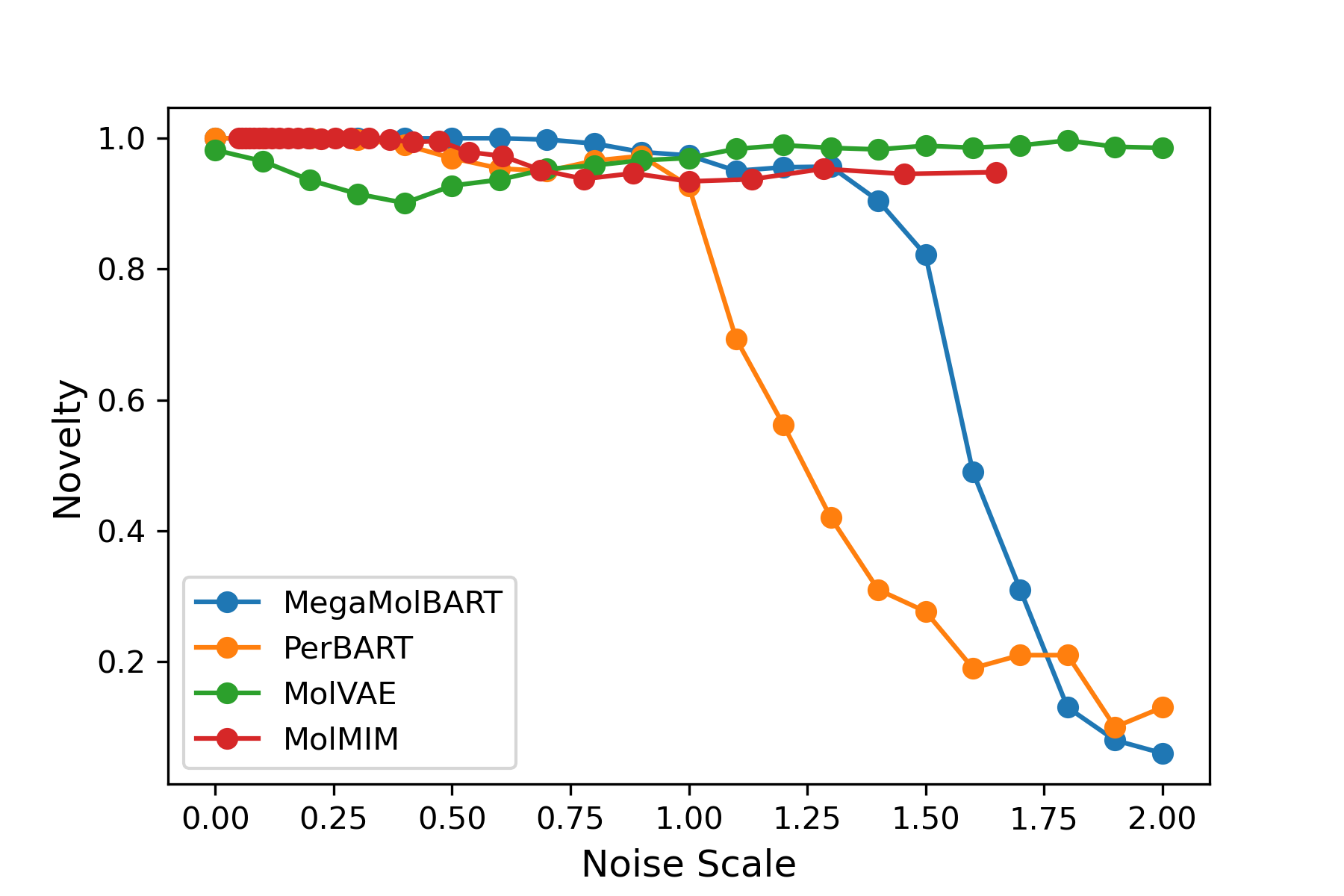}
 \caption{Novelty as a function of noise scale. Note how latent variable models, MolVAE and MolMIM, are consistent while the others, MegaMolBART and PerBART, see a sharp decline due to the increased validity issues at large noise scales.}
  \label{fig:novelty}
\end{figure}

\begin{figure}[htp]
 \centering
 \includegraphics[width=\columnwidth]{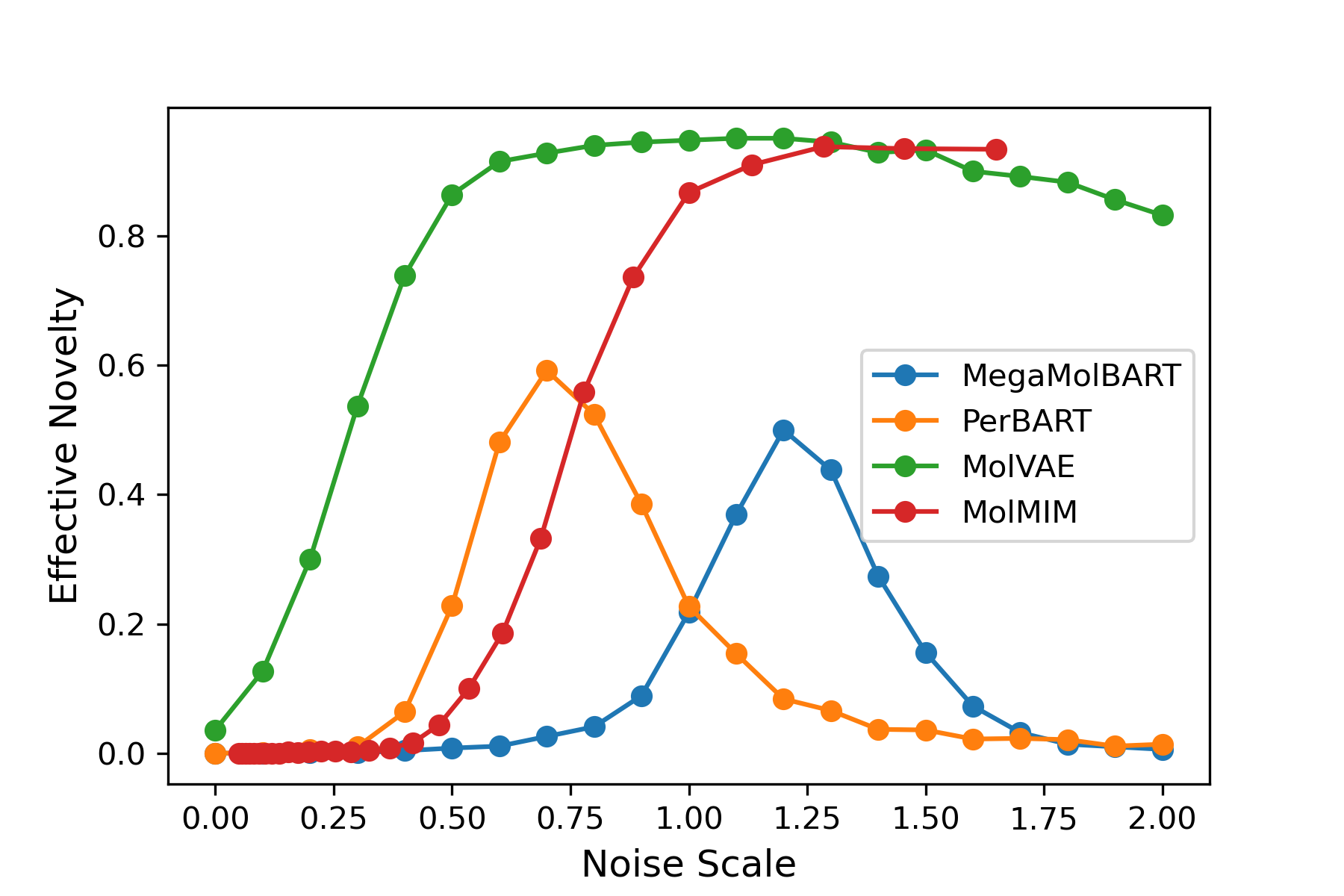}
 \caption{Effective Novelty as a function of noise scale. Note the non-latent variable models, MegaMolBART and PerBART, have a parabolic shape due to validity issues at large noise scales.}
  \label{fig:effective-novelty}
\end{figure}

In this section, we show the results of the hyperparameter search for the optimal noise scale that maximizes effective novelty, per model.
See Figs. \ref{fig:validity}, \ref{fig:unique}, \ref{fig:non-iden}, \ref{fig:novelty}, \ref{fig:effective-novelty}. 
We omit here the search for the optimal hidden length $K$ per model, where we considered $K \in {1, 2, 4, 8, 16}$ for all models.
\FloatBarrier


\subsection{Compute Limited Single Property Optimization} \label{sec:compute-limited}

\begin{table}[H]
\renewcommand{\arraystretch}{1.1}
\centering
\begin{tabular}{l||cccc|cc} 
 & \multicolumn{4}{c|}{\textbf{Success \% of QED (iter)}} & \multicolumn{2}{c}{\textbf{Pen. logP ~($\boldsymbol{\delta=0.4}$})} \\ 
\textbf{Model} & \textbf{100} & \textbf{300} &  \textbf{400} & \textbf{800} & \textbf{avg. $\Delta$} & \textbf{\% ($\Delta > 0$)} \\ 
\hline
\hline
{PerBART} & 2 & 2.12 & - & - & 2.6 ± 2.3 &  23 \\ 
{MolVAE} & 6.6 & 21.2 & - & - & 3.0 ± 2.8 &  40.6 \\ 
\emrow {MolMIM} & \textbf{37} & \textbf{58} & \textbf{66.8}  & \textbf{70.5} & \textbf{4.2 ± 1.6} &  \textbf{78} \\ 
\hline
CDDD & 16 & 38 & 51.0  & 70.2 & 2.1 ± 2.4 &  45 \\ 
\end{tabular}%
\caption{Results based on compute limited to a single restart and specified number of iterations. 
Penalized logP uses 100 iterations. 
We see that MolMIM improves upon all tested methods in terms of success rate for QED, and average increase and improvement percentage for logP. Missing results are due to excessively long run times. Top models are developed herein.}
\label{table:compute-limited}
\end{table}

As a final single property experiment, we explore the above tasks using significantly reduced query budgets, and only a single restart.
We consider 100, 300, 400, and 800 iterations for the QED task, and 100 iterations for the penalized logP task.
Table. \ref{table:compute-limited} shows the results, where MolMIM consistently provides superior results over all other models, 
including CDDD, which is trained with chemical property information. It is important to mention that only CDDD exhibited the generation of invalid molecules during the optimization procedure.
We note that the improved performance of both MolMIM and MolVAE, relative to PerBART demonstrates the importance of having a regularized latent space.
We also point the reader to the significant difference between MolMIM and MolVAE, demonstrating the importance of the learned latent spaces.

\FloatBarrier
\clearpage

\subsection{Sampling Metrics}  \label{sec:sampling-metrics}


\subsubsection{Sampling Metric Formulation}

Here, we formulate the sampling metrics as described in the main body:
\begin{alignat}{3}
&\text{validity} & ~ = ~ & \frac{|V|}{|G|}  \label{eq:valid}  \\ 
&\text{uniqueness} & ~ = ~ & \frac{|U|}{|V|}  \label{eq:uniq}  \\ 
&\text{novelty} & ~ = ~ & \frac{|N|}{|U|}  \label{eq:nov}  \\ 
&\text{non identicality} & ~ = ~ & \frac{|\bar{I}|}{|V|}  \label{eq:ni}  \\ 
&\text{effective novelty} & ~ = ~ & \frac{|N\cap \bar{I}|}{|G|}  \label{eq:enov} 
\end{alignat}
where 
\begin{itemize}
    \item $G$ is the set of all generated molecules
    \item $V$ is the subset of all valid molecules in $G$
    \item $U$ is the subset of all unique molecules in $V$
    \item $N$ is the subset of all novel molecules in $U$
    \item $\bar{I}$ is the subset of all non identical molecules in $V$
\end{itemize}
are the corresponding sets.

The design flaw in Eq. \ref{eq:nov} is that $N$ is a subset of $U$ and therefore does not consider the total amount of generated molecules $G$. 
Effective novelty not only measures the percentage of useful molecules but it also provides a measurement for sampling efficiency as it is defined over all generated molecules in Eq. \ref{eq:enov}.

\FloatBarrier


\subsubsection{Visualizing Effective Novelty}

Novelty is based on molecules that are not present in the training set. However, it does not discriminate against duplicates.
In such a case, one might need to sample multiple times to achieve the desired quantity of novel molecules.
For this reason, it can be convenient to have a single metric that describes the sampling efficiency of the model.
As an example, imagine a model reconstructs the novel input molecule 50\% of the time, and the other 50\% of the time generates different molecules which are valid, novel, and unique.
If such a model produces 10 samples, the novelty will be 100\%, and the uniqueness will be 60\%. 
If the objective is to generate 10 novel molecules, we will have to sample 20 times from this model, which is deceiving for a novelty of 100\%.
To address the above issue and in order to simplify the evaluation of the sampling quality of a model, we introduce two new metrics Non-Identicality and Effective Novelty.
In the case of the example above, the effective novelty will be 50\%.

\begin{figure}[htp]
\centering
\includegraphics[width=0.9\columnwidth]{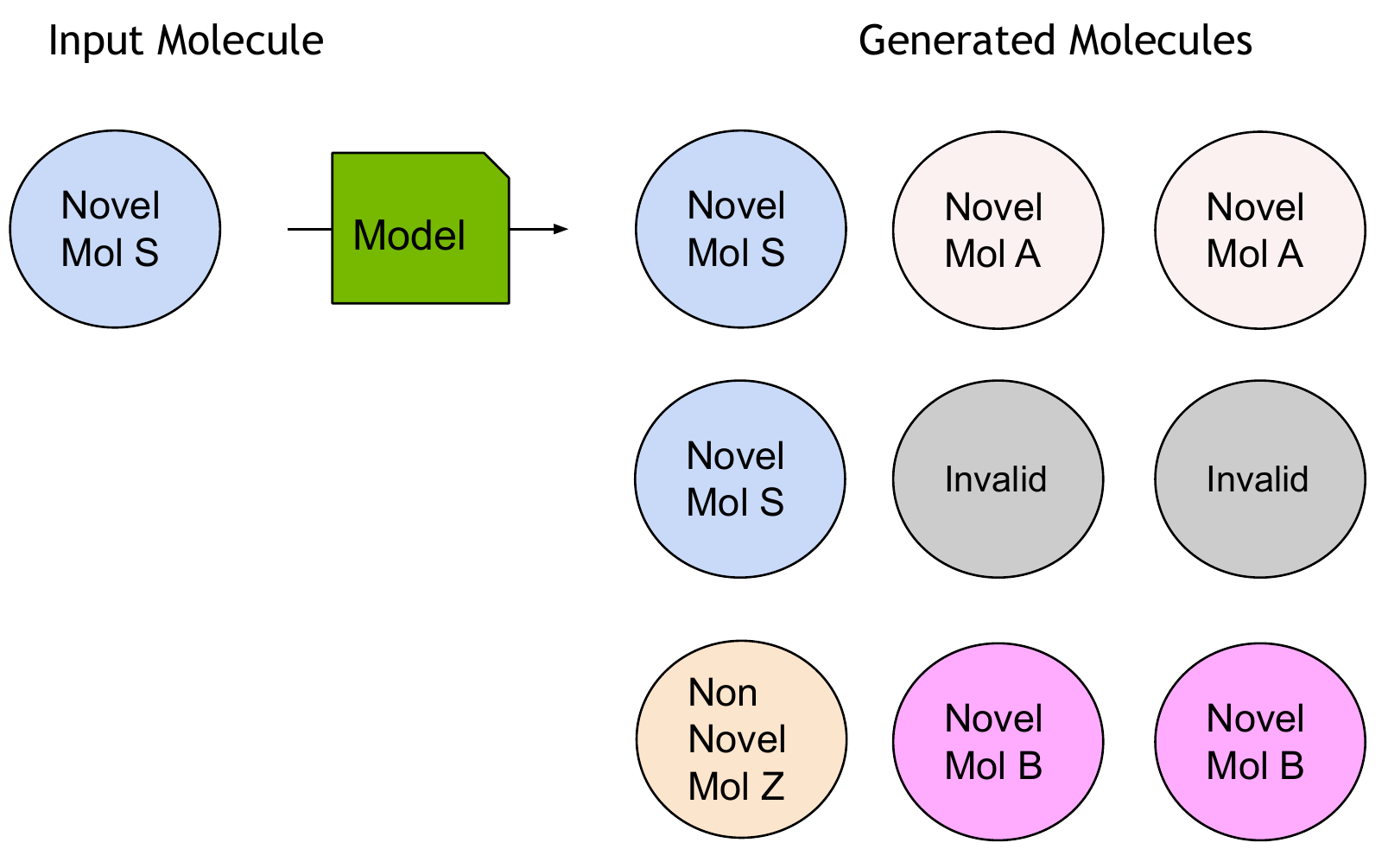}
\caption{Given a novel input molecule S, we provide an example sampling output used in Figure. \ref{fig:effective-novely-right} to visualize all of the defined metrics Eq. \ref{eq:valid} - \ref{eq:enov}}
\label{fig:effective-novely-left}
\end{figure}

\begin{figure}[htp]
\centering
\includegraphics[width=0.9\columnwidth]{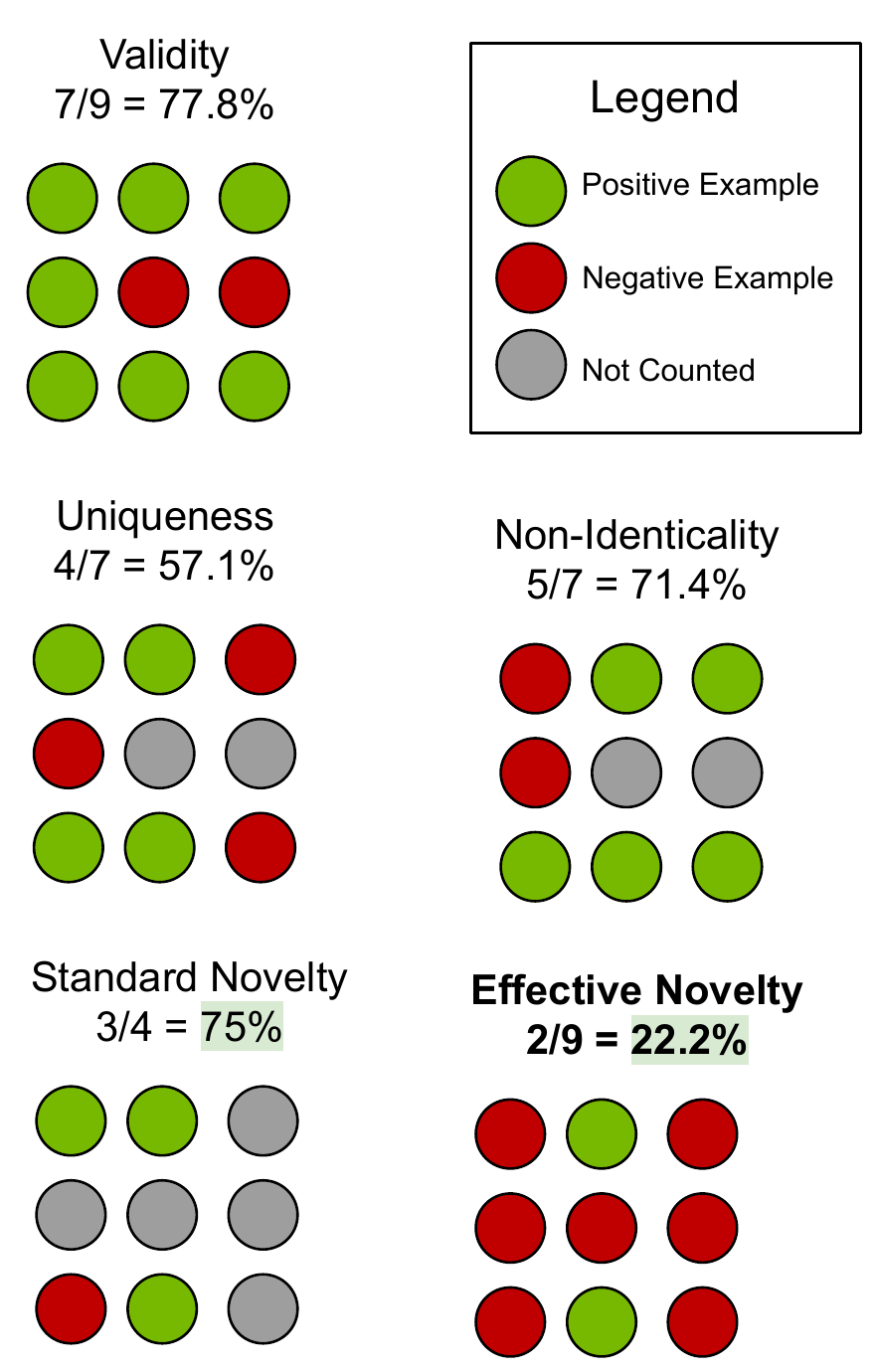}
\caption{Here, we highlight the difference between novelty and effective novelty. For each metric, we mark the numerator (green), the denominator (green + red), and the irrelevant part(gray).
The high novelty might still lead to low efficiency in sampling novel molecules.}
\label{fig:effective-novely-right}
\end{figure}

We provide a depiction of effective novelty in Figs. \ref{fig:effective-novely-left}-\ref{fig:effective-novely-right} for further understandings..

\end{document}